 \documentclass[final,5p,times,twocolumn,authoryear]{elsarticle}

\usepackage{amssymb}
\usepackage{lipsum}
\usepackage{hyperref}
\hypersetup{
    colorlinks=true,
    linkcolor=blue,
    filecolor=magenta,      
    urlcolor=cyan,
    pdftitle={Overleaf Example},
    pdfpagemode=FullScreen,
    }
\usepackage{subcaption}
\usepackage{enumitem}

\begin{document}

\begin{frontmatter}



\title{XAI for All: Can Large Language Models Simplify Explainable AI?}


\author[first]{Philip Mavrepis}
\author[first]{Georgios Makridis}
\author[first]{Georgios Fatouros}
\author[first]{Vasileios Koukos}
\author[fifth]{Maria Margarita Separdani}
\author[first]{Dimosthenis Kyriazis}

\affiliation[first]{organization={University of Piraeus, Department of Digital Systems},
           addressline={Karaoli ke Dimitriou 80}, 
           city={Piraeus},
           postcode={18534}, 
           state={Attica},
           country={Greece}}
\affiliation[fifth]{organization={University of Piraeus, Department of Maritime Studies},
           addressline={Karaoli ke Dimitriou 80}, 
           city={Piraeus},
           postcode={18534}, 
           state={Attica},
           country={Greece}}

\begin{abstract}
The field of Explainable Artificial Intelligence (XAI) often focuses on users with a strong technical background, making it challenging for non-experts to understand XAI methods. This paper presents "x-[plAIn]", a new approach to make XAI more accessible to a wider audience through a custom Large Language Model (LLM), developed using ChatGPT Builder. Our goal was to design a model that can generate clear, concise summaries of various XAI methods, tailored for different audiences, including business professionals and academics. The key feature of our model is its ability to adapt explanations to match each audience group's knowledge level and interests.  Our approach still offers timely insights, facilitating the decision-making process by the end users. Results from our use-case studies show that our model is effective in providing easy-to-understand, audience-specific explanations, regardless of the XAI method used. This adaptability improves the accessibility of XAI, bridging the gap between complex AI technologies and their practical applications. Our findings indicate a promising direction for LLMs in making advanced AI concepts more accessible to a diverse range of users.
\end{abstract}



\begin{keyword}
Explainable AI \sep Human-Centric Explainable AI \sep LLM \sep GPT Builder \sep Audience Analysis \sep XAI \sep AI
\end{keyword}

\end{frontmatter}




\section{Introduction}
\label{introduction}

In the contemporary epoch, frequently denoted as the Digital or Information Age, a characteristic feature is the proliferation of sophisticated computational systems generating copious data on a daily basis. This epoch is further defined by the digital metamorphosis occurring within industrial realms, culminating in the advent of the fourth industrial revolution, Industry 4.0 \cite{makridis2020predictive}. The cornerstone of this revolutionary phase is AI, which stands as the pivotal facilitator of the Industry 4.0 paradigm, fostering the development of innovative tools and processes \cite{soldatos2021trusted}. Simultaneously, there is an escalating intrigue in XAI, which is oriented towards providing intelligible explanations for the inferences and choices formulated by machine learning algorithms.

The pivotal contribution is based on an innovative paradigm in the realm of human-centric XAI. This paper introduces a ground-breaking GPT-based LLM, serving as a versatile interface that empowers end-users to intuitively comprehend and interpret results derived from a multitude of XAI methodologies. This model is characterized by its:

\begin{enumerate}
    \item Audience-Adaptive Explanations: The core achievement lies in its capability to produce concise, easily digestible summaries of complex XAI methods, specifically tailored to align with the varying expertise levels and interests of diverse audience groups, ranging from business professionals to academic researchers. This customization enhances user engagement and understanding across different sectors.
    
    \item XAI Methodology Agnosticism: A unique attribute of this model is its agnostic approach to XAI methods. This design ensures broad applicability and relevance across a wide spectrum of XAI techniques and knowledge domains, without necessitating specific training or adaptation for each distinct method. This flexibility marks a significant advancement in the field of XAI.
    
    \item Decision-Making Facilitation: The model's capacity to provide timely, clear, and contextually relevant explanations significantly augments decision-making processes for end-users. This aspect is particularly crucial in scenarios where comprehension of AI outputs is essential for critical decision-making but is hindered by the technical complexity of XAI outputs.
    
    \item Empirical Validation through Use-Case Studies: The practical efficacy of this LLM is further underscored by empirical evidence gathered from use-case studies. These studies demonstrate the model's effectiveness in delivering audience-specific explanations that are comprehensible and relevant, thereby validating the model's applicability and impact in real-world scenarios.
\end{enumerate}

In essence, this paper propels the field of XAI towards greater inclusivity and practicality, by innovatively merging advanced AI concepts with user-friendly interfaces. This approach not only demystifies XAI for non-experts but also significantly contributes to the broader adoption and understanding of AI technologies in various professional contexts.

The remainder of the paper is organized as follows: Section 2 presents the background and the motivation of our research, while Section 3 delivers the literature review in the areas of study of this paper. Section 4, presents the proposed methodological approach, introduces the overall implementation, and offers details regarding the datasets used and the evaluation procedure. Section 6 dives deeper into the results of the conducted research and the corresponding survey. Section 5 concludes with recommendations for future research and the potential of the current study.

\section{Background}
\par Our research's underlying motivation is illuminated through an introduction to the foundational concepts of Image Classification, XAI methods, and adversarial attacks.

\subsection{eXplainable AI (XAI)}
\par Interpretability or explainability in Machine Learning (ML) models refers to the ability to describe and understand an ML model's workings \cite{choo2018visual}. This is particularly vital in Deep Neural Networks (DNN), which are inherently complex and thus perceived as "black boxes" \cite{zahavy2016graying}. The burgeoning field of research addressing the opacity of these ML "black boxes" is known as XAI \cite{gunning2016explainable}.

\par
\begin{figure*}[h!]
\centering
\includegraphics{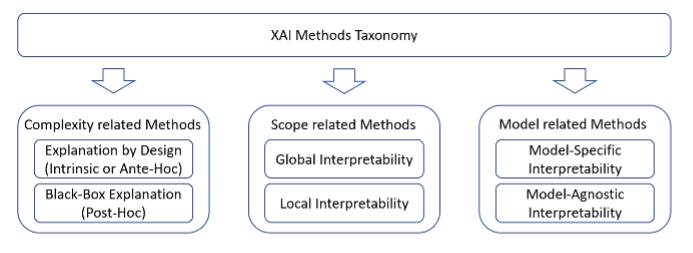}
\caption{Taxonomy of XAI Methods}
\label{fig:taxonomy}
\end{figure*}

 Herein, XAI assumes a critical yet sensitive role, acting as a conduit between intricate DL models and those without IT expertise. Consequently, XAI methodologies must be precise and comprehendible to domain experts, fostering a sense of "trust" in real-time settings. Over the past few years, several XAI methods, strategies, and frameworks have emerged. For our research, we categorize XAI methods based on their simplicity, the degree of interpretability, and the dependency level on the analyzed ML/AI model, as illustrated in Figure \ref{fig:taxonomy}.

\par Moreover, complexity-related methods in XAI can be bifurcated into i) intrinsically explainable (Ante-Hoc) models, also known as transparent or glass box approaches, and ii) black-box (Post-hoc) models, which necessitate deciphering the reasoning steps behind predictions for explainability purposes. Additionally, these methods can be categorized based on their scope: i) global explainability methods, which scrutinize the algorithm as a whole, including training data and proper algorithm usage, and ii) Local explainability, which pertains to the system's ability to elucidate specific decision-making processes.

Lastly, it's crucial to differentiate between model-specific and model-agnostic XAI approaches. The key difference lies in whether the XAI method depends on the underlying ML model or if it can be universally applied.


\subsection{From explainability to interpretability}
In scholarly discussions, a notable discrepancy persists regarding the precise definitions of "explainability" and "interpretability." While these terms are often used interchangeably, some scholars distinguish between them, as noted in \cite{arrieta2020explainable} and \cite{chakraborty2017interpretability}.

This analysis adheres to the differentiation between explainability and interpretability as explicated in \cite{saeed2023explainable}. According to this reference, explainability entails the provision of insights tailored to satisfy a specific requirement of a designated audience, while interpretability is concerned with the extent to which these insights are comprehensible and relevant within the framework of the audience’s specialized knowledge base.

Explainability is defined by three fundamental elements, as outlined in the aforementioned source: the nature of the insights provided, the specific audience targeted, and the underlying necessity for these insights. These insights emanate from various methodologies in explainability, such as textual descriptions, the significance of features, or localized elucidations, and are intended for a diverse audience including sector-specific professionals, individuals directly impacted by the outcomes of the model, and experts in the field of model development.

In the realm of interpretability, the focus shifts to the congruence and logicality of the explanations about the targeted audience's pre-existing knowledge base. This includes assessing whether the explanations are coherent and meaningful to the audience, if the audience is capable of employing these explanations in their decision-making processes, and whether the explanations provided offer a rational basis for the decisions made by the model.

\subsection{Challenges in Communicating AI Concepts}
Communicating the concepts of AI to a broad audience encompasses a multitude of challenges, stemming from the inherently complex and rapidly evolving nature of AI technology. These challenges are amplified when discussing the domain of XAI, where the goal is to make AI decision-making processes transparent and understandable to various stakeholders.

Although the various available open-source XAI algorithms, such as  LIME (Local Interpretable Model-agnostic Explanations) \cite{ribeiro2016should}, SHAP (SHapley Additive exPlanations) \cite{lundberg2017unified}, and Gradient-weighted Class Activation Mapping (Grad-Cam) model \cite{selvaraju2017grad}, examples of XAI in real-world applications still need to be discovered. The root cause of this is that SotA XAI algorithms aim to assist the developer of the AI system instead of the end-user. Developing XAI applications needs human-centered
approaches that align technical development with people's explainability needs and define success by human experience, empowerment, and trust. Furthermore, AI algorithms can exhibit various forms of bias \cite{klein2020reducing}, including social, racial, and gender prejudices. XAI and Exploratory Data Analysis \cite{torralba2011unbiased}.
However, implementing bias mitigation and XAI techniques in a larger situational context (i.e., explaining multiple AI models that perform a single task) becomes increasingly more complicated. Cutting-edge XAI approaches are rigorously disconnected, with just a local input view linked with each particular AI model utilized throughout the overall (global) reasoning process \cite{jan2020ai}. Moreover, existing techniques usually lack reasoning semantics and remain detached from the broader process context.

Other challenges rely significantly on the utilized interface between the human and machine/software. An effective HMI should consider various aspects such as the level of autonomy, user expertise, use case/domain \cite{lim2009assessing}, as well as security and trust \cite{virtue2017designing}. Despite the extended research, many works suggest that designers need more guidance in designing interfaces for intelligent systems [\cite{baxter2018meet} that could be used by the non-IT-savvy public.

\subsection{Objectives}
This paper aims to operationalize human-centered perspectives in XAI at the conceptual, methodological, and technical levels toward Human-Centred Explainable AI (HC-XAI) models. We enhance cutting-edge XAI approaches for explaining ML models, and models that explain deep neural networks to HC-XAI models, shaping the final output of black-box models considering the context and biases while allowing feedback and adjustment from the user.

Our research was motivated based on two extensive surveys to explore the challenges and preferences in the field of XAI, and they are focusing on two challenging areas considering XAI modeling. These two areas are the Time-series Classification \cite{makridis2023xai} and Vessel Route Forecasting (VRF) \cite{makridis2023towards}. Both of them are quite challenging in terms of AI interpretability for various reasons such as the complexity of quantifying explainability in XAI, highlighting the subjective nature of explainability and the diverse range of stakeholders involved. The results highlight the inherent subjectivity in explainability, with different individuals having varied preferences and understandings. This underscores the need for a flexible approach in designing explainability communication to end users (Human-centric XAI). However, a strong preference for visualization techniques was revealed, such as overlaying predicted versus actual trajectories, indicating the importance of visual methods in making explanations understandable. Given the responses on both surveys, it is clear that a description of the visualizations as a complementary tool is desired by non-IT end users. 

Based on these findings, we propose an integrative approach that combines the strengths of visual and textual explanations. This approach aims to make XAI results more human-centered, focusing on providing user-friendly interfaces such as chatbox-based human-AI interactions, ensuring that the design of explanations and interfaces is user-centric, focusing on the specific needs and preferences of different user groups. This also involves propositions of the decision-making process to offer added values based on the XAI outcomes. This approach not only aims at demystifying AI decisions but also at enriching the user's understanding by providing context-rich, detailed insights into the AI's decision-making process.

\section{Literature Review}

\subsection{Explainable AI: A Technical Overview}
As complex predictive models are increasingly integrated into areas traditionally governed by human judgement, there is a growing demand for these models to offer more clarity in how they reach decisions \cite{susnjak2023beyond}. This transparency is vital for building trust and meeting regulatory compliance, especially in international legal contexts where explaining automated decisions affecting people is becoming a legal necessity. According to \cite{wachter2017counterfactual}, it's also crucial that individuals can challenge decisions made by these systems and understand what changes in their data could lead to different outcomes. Technologies like Counterfactuals have been developed to provide insights into minimal changes needed for different predictions by these models.

This need for clarity has given rise to the field of XAI or Interpretable Machine Learning. This area aims to create methods that make complex predictive models more understandable and tools that explain how these models formulate their conclusions (\cite{molnar2020interpretable}). Additionally, there's growing interest in prescriptive analytics, which focuses on using data to create actionable insights \cite{lepenioti2020prescriptive}.

From a technical standpoint, model interpretability involves understanding the internal workings of a machine learning model post-training, generally at a broad, global level. Conversely, model explainability delves into understanding the rationale behind a model's prediction for a specific instance, known as local-level explainability. Both are important: interpretability allows institutions to broadly explain how a model works to stakeholders, while local-level explainability facilitates validating specific predictions and providing detailed feedback to those affected, like students identified as at-risk.

In the pursuit of model transparency, tools like SHAP, recognized as a leading visualization technique in XAI, provide insight into both global and local-level transparency \cite{gramegna2021shap}. The Anchors technique (\cite{ribeiro2018anchors}) offers a high degree of local-level explainability through human-readable, rule-based models. Furthermore, advanced Counterfactuals not only enhance predictive analysis but also enable prescriptive suggestions, helping learners understand the changes needed for a different outcome. This study showcases the application of these technologies across various stages of the proposed prescriptive analytics framework.

\subsection{Language Models in AI}

Following the success of GPT, a range of LLMs have been developed, exhibiting impressive capabilities in various Natural Language Processing (NLP) tasks, including those in finance.

One standout model in this domain is BloombergGPT, created by Bloomberg's AI team and trained on an extensive collection of financial texts. It has shown exceptional proficiency in financial NLP tasks \citep{wu2023bloomberggpt}. However, as of May 2023, BloombergGPT remains largely for internal use at Bloomberg, lacking a publicly accessible API.

Google's Bard, a key competitor to ChatGPT, is another notable LLM. Powered by Google’s LAMDA (Language Model for Dialogue Applications), it merges aspects of BERT and GPT to facilitate engaging, contextually aware conversations \citep{thoppilan2022lamda}. Like BloombergGPT, Bard also doesn't offer an open API as of this writing.

BLOOM, an open-source contender to GPT-3 \citep{scao2022bloom}, has also gained attention in the LLM space. While it's open-source, effectively using BLOOM requires considerable technical know-how and computing power, and it lacks a version fine-tuned for conversational tasks, a feature where models like ChatGPT excel.

Since ChatGPT's introduction, numerous LLMs have emerged targeting specific functions, such as code completion \citep{dakhel2023github}, content generation, and marketing. These models offer specialized utility, expanding the scope and impact of LLMs. ChatGPT continues to lead in the field \citep{jasper23report}, thanks to its open API, extensive training data, and versatility across various tasks. Despite ChatGPT's broad application in fields like healthcare and education \citep{sallam2023chatgpt}, its direct use in financial sentiment analysis is relatively uncharted. \cite{fatouros2023transforming} presents evidence that ChatGPT, even when applied with zero-shot prompting, can understand complex contexts requiring advanced reasoning capabilities. In addition, MarketSense-AI, a real-world financial application, leverages GPT-4 with Chain-of-Thought (CoT) to effectively explain investment decisions \cite{fatouros2024can}.

\subsection{Large Language Models in XAI}

Significant advancements have been made in AI and LLMs based on transformers, which now exhibit near-human proficiency in text generation and discourse. This progress is largely attributed to their ability to understand long-range dependencies and contextual nuances in texts, thanks to self-attention mechanisms. Models like Google's BERT (\cite{devlin2018bert}) and OpenAI's latest GPT series have set new benchmarks in various natural language processing tasks, including text generation (\cite{brown2020language}). OpenAI's most recent development, the ChatGPT model, exemplifies these advancements by effectively translating complex analytical outputs into user-friendly, actionable language, aiding learners and advisors.

In the realm of cybersecurity, HuntGPT utilizes the capabilities of LLMs and XAI to enhance network anomaly detection. It integrates a Random Forest classifier with the KDD99 dataset \cite{misckddcup1999data130}, advanced XAI frameworks, and the power of GPT-3.5 Turbo. HuntGPT not only detects threats with remarkable accuracy but also conveys them in a clear, understandable format, greatly improving decision-making for cybersecurity experts \cite{ali2023huntgpt}. While, \cite{chun2023explainable} delves into the fusion of XAI with Computational Digital Humanities. It investigates diachronic text sentiment analysis and narrative generation using advanced LLMs like GPT-4. Additionally, it introduces an innovative XAI grey box ensemble. This ensemble combines top-tier model performance with superior interpretability and privacy, underpinned by novel local and global XAI metrics.

\section{Methodology}

Initially, an assessment of state-of-the-art XAI techniques, such as LIME, SHAP, and GradCam and PDP. These methods will be adapted and integrated into the customized LLM infrastructure, focusing on generating natural language explanations delivered via the AI Chat Interface. This integration aims to transform complex XAI visualizations into user-friendly narratives and insights, interpretable by end users.

\subsection{Role of GPT-Builder in LLM Development}

The development of LLMs such as GPT variants has revolutionized the field of natural language processing (NLP). A critical component in this evolution is the role of tools like GPT-Builder, a sophisticated framework for constructing, fine-tuning, and deploying these advanced models. GPT-Builder serves as a pivotal element in LLM development, offering a blend of user-friendly interfaces and powerful backend processes that streamline the creation and management of these complex models \href{https://chat.openai.com/gpts/editor}{GPT Builder}.

GPT-Builder plays an instrumental role in democratizing access to LLM technology. It empowers organizations and individual developers to build custom LLMs tailored to specific needs or domains. This customization is crucial in scenarios where a standard GPT model may not provide optimal performance, such as in specialized professional fields or for languages and dialects with limited representation in mainstream models. GPT-Builder simplifies the process of training these models on niche datasets, making it feasible for non-experts in machine learning to develop highly specialized and effective LLMs.

\subsection{Use cases}\label{sec:usecases}
In our study, we developed a custom GPT model, the x-[plAIn] GPT. x-[plAIn] model underwent extensive testing across a diverse range of XAI methods and problem definitions. However, for the interactive component involving end users, we focused on five specific use cases presented through a questionnaire. We made a concerted effort to select XAI implementations that spanned various sectors and catered to different levels of technical expertise.

\subsubsection{Use Case 1}
The first use-case featured in our study was derived from \cite{makridis2022evaluating} which investigated the detection of boar taint. In this research, the authors identified significant factors contributing to the boar-taint phenomenon, employing SHAP values among other methods. This particular implementation of SHAP values was incorporated into our questionnaire.

\begin{figure}[ht!]
    \centering
    \includegraphics[width=1\linewidth]{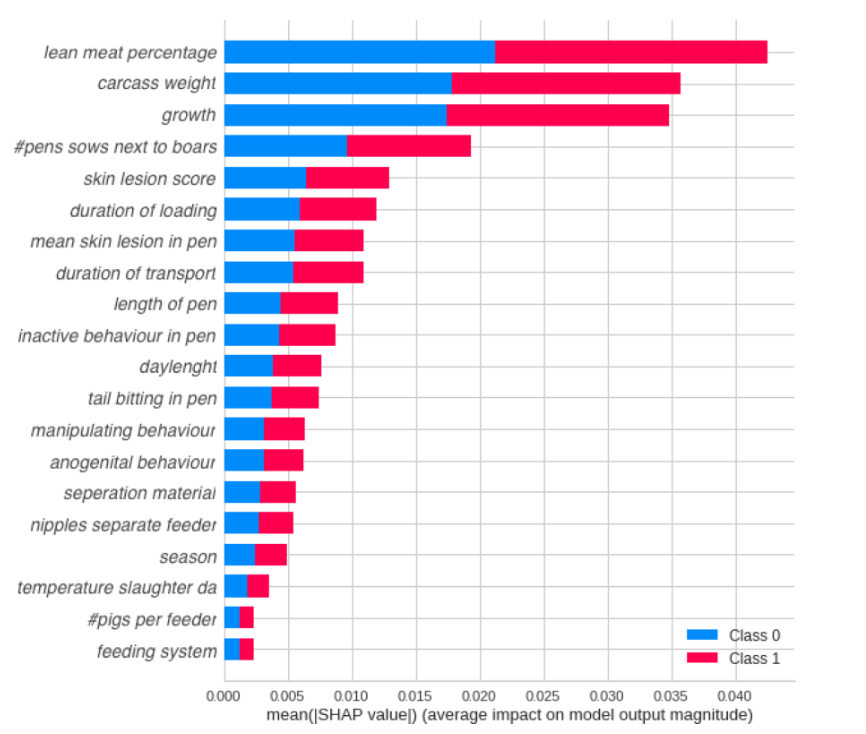}
    \caption{Plot for SHAPley values evaluation in \cite{makridis2022evaluating}.}
    \label{fig:xplain_role}
\end{figure}

\subsubsection{Use Case 2}
The second use-case was based on \cite{szczepanski2021new} where the authors explored the use of LIME and Anchors (XAI methods) for generating explainable visualizations in the context of fake news detection. This study represented another facet of XAI application, showcasing its utility in media and information analysis.

\begin{figure}[ht!]
    \centering
    \includegraphics[width=1\linewidth]{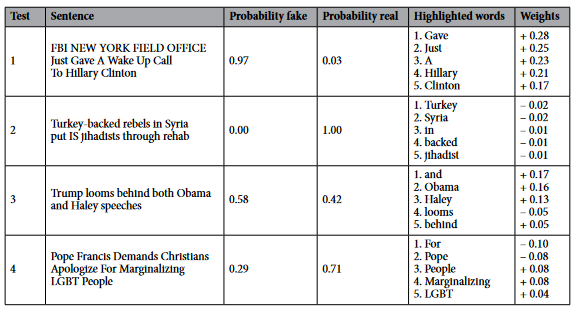}
    \caption{Plot for LIME evaluation of fake news from \cite{szczepanski2021new}.}
    \label{fig:xplain_role}
\end{figure}

\subsubsection{Use Case 3}
The third use-case in our study involved the visualizations developed by \cite{feldhus2023saliency}. The authors employed the Integrated Gradients feature attribution method to represent the predictions made by a BERT model. Building on this, they created a model-free and instructed (GPT-3.5) Saliency Map Verbalization (SMV) explaining the prediction representations.

\begin{figure}[ht!]
    \centering
    \includegraphics[width=1\linewidth]{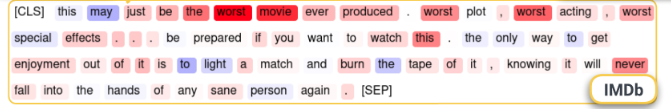}
    \caption{Plot for Saliency Map Verbalization (SMV) from \cite{feldhus2023saliency}.}
    \label{fig:xplain_role}
\end{figure}

\subsubsection{Use Case 4}
The fourth application incorporated XAI techniques as implemented by \cite{moujahid2022combining}. In their study, Grad-CAM was employed to identify regions of interest pertinent to the prediction of COVID-19 in lung X-ray images, utilizing various network architectures.

\begin{figure}[ht!]
    \centering
    \includegraphics[width=1\linewidth]{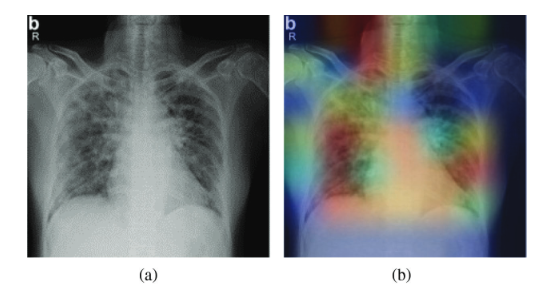}
    \caption{Plot for Grad-CAM explanations from \cite{moujahid2022combining}.}
    \label{fig:xplain_role}
\end{figure}

\subsubsection{Use Case 5}
Lastly, the fifth use-case presented substantial technical complexities. The researchers in \cite{moosbauer2021explaining} employed Partial Dependence Plots (PDP), an infrequently used method within XAI, for the purpose of hyperparameter optimization. Consequently, they generated and scrutinized plots to exhibit robust and trustworthy Partial Dependence (PD) estimates across an intelligible subset of the hyperparameter space, considering a variety of model parameters.

\begin{figure}[ht!]
    \centering
    \includegraphics[width=1\linewidth]{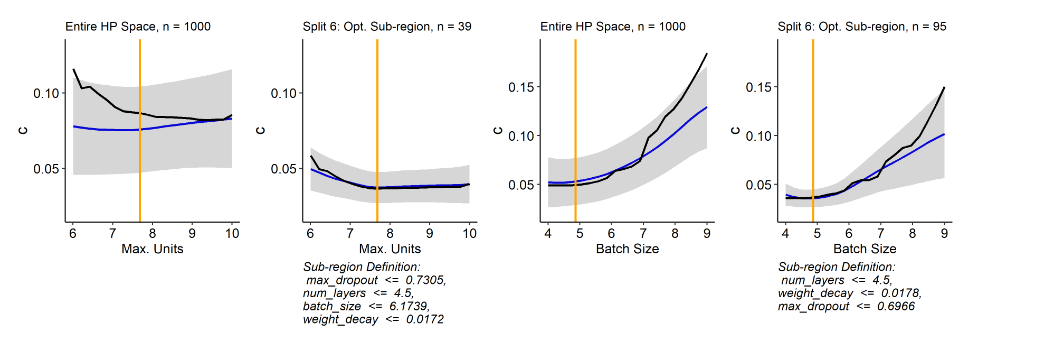}
    \caption{Plot for Partial Dependence Plots (PDP) from \cite{moosbauer2021explaining}.}
    \label{fig:xplain_role}
\end{figure}

\subsection{Baseline: XAI approaches}

Our innovative approach in XAI integrates outputs from LIME, SHAP, and Grad-CAM into a GPT masses enhanced textual and data analysis from these tools.

\begin{itemize}
    \item LIME interprets complex models by approximating them locally with simpler models, revealing feature influence on predictions. It's insightful but limited by potential instability and local focus.

    \item SHAP, rooted in game theory, assesses feature importance globally, offering consistent and fair interpretations but at the cost of computational complexity and possible non-intuitiveness.

    \item GradCam identifies important image regions for predictions, improving versatility and accuracy while maintaining intuitiveness.
\end{itemize}

\subsection{LLM Enhanced XAI explainer}

In the development of our GPT-based XAI explainer, a rigorous, technical approach was employed. Initially, we utilized a specialized interface for defining the explainer's core objectives, focusing on advanced interpretability for AI decision-making processes. The configuration phase was comprehensive, involving precise customization of the model's parameters, including its naming, operational descriptions, and initialization prompts tailored for nuanced AI explanations. The development of the prompt engineering process is delineated in Table \ref{tab:xai_prompts}. This progression adheres to the prompt-engineering guidelines provided in the official documentation of OpenAI, accessible via this \href{https://platform.openai.com/docs/guides/prompt-engineering}{hyperlink}. Additionally, acknowledging the paramount significance of causal explanations and insight generation for HC-XAI, we incorporated a CoT approach into our final prompt. Given the absence of universally correct responses for this task, we eschewed the methodology detailed in \cite{wei2022chain}. Instead, we adopted a more straightforward strategy, integrating the phrase "Let's think step by step." This inclusion has proven to be notably effective, as substantiated by \cite{tan2023causal}. This technical methodology ensured the creation of a GPT model specifically fine-tuned for the complexities and demands of XAI, enhancing its effectiveness in delivering clear, understandable insights into AI decisions.

\begin{table*}[htbp]
\scriptsize
\centering
\caption{Gradual Development of ChatGPT Prompts for XAI Method Simplification}
\label{tab:xai_prompts}
\begin{tabular}{p{1.0cm}p{11.0cm}p{6cm}}
\hline
\textbf{Prompt Version} & \textbf{Prompt} & \textbf{Benefit Over Previous Version} \\
\hline
P1 & Provide summaries of insights from XAI methods, focusing on clarity and relevance. Avoid including obvious elements unless specifically requested. & Introduces the basic concept of summarizing XAI insights with a focus on clarity and relevance. \\
\hline
P2 & Summarize insights from XAI methods like LIME and SHAP. Focus on clarity and relevance, and begin to consider the user's context in your summaries. Exclude obvious elements unless they are explicitly asked for. & Specifies XAI methods (LIME and SHAP) and introduces the concept of tailoring summaries to the user's context. \\
\hline
P3 & Generate clear and relevant summaries from XAI methods such as LIME and SHAP, tailored to the user's context. Begin to integrate actionability into the insights and ask the user for their expertise level (beginner, intermediate) before responding. & Adds the element of actionability and the need to adjust responses based on the user's expertise level. \\
\hline
P4 & Provide clear, relevant, and actionable summaries of insights from XAI methods like LIME and SHAP. Tailor the content to the user's expertise level and specific inquiries. Include practical suggestions or conclusions and avoid obvious elements from the input unless requested. & Emphasizes the customization of summaries to specific user inquiries and the inclusion of practical suggestions or conclusions. \\
\hline
P5 & Objective: Deliver concise summaries of insights from XAI methods like LIME and SHAP, tailored to user's context and expertise level. Focus on clarity, relevance, actionability, and responsiveness. Ask the user for their expertise level and tailor your response accordingly. Avoid including obvious input elements unless explicitly asked. Provide practical suggestions or conclusions. & Introduces the concept of responsiveness to user-specific inquiries and further emphasizes tailoring content based on expertise. \\
\hline
P6 & Objective: Provide concise, user-friendly summaries of insights derived from XAI methods. 
If multiple insights can be drawn from a single input try to combine them into a larger context. Let's think step by step on how the final insight is reached.
\newline\newline
Output Expectations:\newline\newline
    Clarity: Deliver straightforward and easily comprehensible summaries.\newline
    Relevance: Ensure insights are directly applicable to the user's context or domain.\newline
    Actionability: Focus on providing practical suggestions or conclusions.\newline
    Responsiveness: Tailor summaries to answer user-specific inquiries based on the XAI analysis.\newline\newline
DO NOT include obvious elements (numbers, text) from the given input unless EXPLICITLY asked.
BEFORE ANSWERING\newline
1. Ask the user for his expertise level (beginner, intermediate).\newline
2. Ask the user's domain of expertise (if none provided assume it aligns with the domain of the input provided)\newline

If the user is intermediate provide information about how he should understand the provided input and then the insight(s).\newline
If the user is beginner DO NOT provide information about the provided input PROVIDE ONLY THE INSIGHT.\newline

Tailor responses to a user's technical and domain expertise.\newline
Provide examples that match the expertise of the user in analogous manner to explain the insights.& Fully integrates all elements including clarity, relevance, actionability, responsiveness, and user-specific customization, creating a comprehensive and detailed approach to summarizing XAI insights. \\
\hline
\end{tabular}
\end{table*}

\subsection{Audience Analysis and Content Customization}
This tool primarily serves two key demographics: end-users of XAI methods and AI developers, notably data scientists, who utilize XAI methods for model understanding. The former category, end-users, typically possesses limited technical knowledge but may exhibit considerable domain expertise. Their primary interest lies in deriving insights from the XAI methods, rather than comprehending the technical intricacies of how these methods operate or the underlying model training processes. Conversely, highly technical users, such as AI developers, leverage this tool to gain deeper insights into their models. They focus on understanding the training mechanisms of the models, identifying potential biases highlighted by the XAI methods, and exploring strategies to address these issues.

The design of this tool is interactive and user-centric, enabling it to evaluate the user's proficiency in AI and XAI methodologies, as well as any domain-specific expertise they may possess. Following this assessment, the tool adeptly tailors its responses, adjusting the focus of its answers to align with the user's knowledge level. This approach ensures that the insights provided are not only relevant and actionable but are also derived effectively from the input given by the user.


\subsection{Evaluation - Feedback}

In our comprehensive study to evaluate the applicability and effectiveness of our GPT-based XAI explainer, we conducted an extensive survey targeting a broad spectrum of professionals. This survey, which can be accessed here, was designed to gather insights into the various aspects of XAI in the context of our GPT model.

Survey Design and Purpose:
The survey was meticulously structured to probe into the respondents' understanding and experiences with AI, Machine Learning (ML), and Deep Learning (DL), as well as their exposure to and perceptions of XAI methods. This allowed us to gauge the baseline knowledge of our audience, which is crucial in tailoring the XAI components of our GPT model.

Assessing User Familiarity and Application of AI:
One of the key objectives was to understand how familiar the respondents were with AI, especially in the context of using AI for specific tasks. This information is vital to ensure that our GPT XAI explainer is accessible to users with varying levels of AI expertise.

Understanding Preferences in Data Description:
The survey extensively examined various applications of XAI methods, including LIME, SHAP, and Grad-CAM, each presented with two distinct descriptions. The first was the original description from the research papers, selectively modified to provide the end user with essential information. In contrast, the second description was generated by the x-[plAIn] GPT model in response to the query, \textit{"What are the top insights from this picture?"} Notably, in instances where the problem definition was not evident or deducible from the input, the model's query included the specific research problem addressed in the original paper. This approach enabled a comprehensive understanding of user preferences regarding textual explanations, including aspects like structure, length, and formality, allowing for subsequent fine-tuning of the model.

\subsection{Limitations}
This tool demonstrates a remarkable ability to interpret various outputs from XAI methods, offering insightful and targeted explanations. However, it has been observed that there are instances in which the model mistakenly attempts to explain the provided image rather than focusing on the XAI output. An illustrative case of this behavior can be found in use case \#3 (SMV), as detailed in Section \ref{sec:usecases}. In this particular example, the model states:

\noindent \textbf{Highlighted High-Impact Negative Phrases:}
\begin{itemize}[label={--}]
    \item The phrases "the worst movie ever produced," "worst plot," "worst acting," and "worst special effects" are strongly emphasized in the saliency map. This implies that these phrases are key elements the model associates with a negative review.
    \item The recommendation to "light a match and burn the tape" is exceedingly negative, indicating a high level of dissatisfaction.
\end{itemize}

While the second observation remains accurate and provides valuable insights, it is not directly related to the actual XAI output since those words are not prominently highlighted by the saliency map.

In general, the model's contextual interpretation can yield generalized insights that may extend beyond the strict confines of the XAI output. Striking the right balance between the precision of the response and the breadth of the insights provided poses a challenge. The most effective approach is to engage an active end-user who employs critical thinking and maintains an open-minded approach to understanding the results.

\section{Results and Discussion}

In our endeavor to delve into the usability and effectiveness of x-[plAIn], we administered a meticulously crafted survey to an eclectic mix of partners and participants, that can be found \href{https://docs.google.com/forms/d/e/1FAIpQLSe20omwn6_RHxJRK_UXnjSHwI8wZyHmQYJ5pmk_HqDo7j_cFg/viewform?usp=sf_link}{here}. Drawing from real-world scenarios, we simulated a context where AI models transition from mere decision-support mechanisms to primary decision-makers, emphasizing their paramount need for transparency and trustworthiness. The survey pivoted on two primary axes: gauging participants' baseline familiarity with AI, ML, and DL; and discerning their perception of key data interpretation based on their experience on AI-enhanced decision-making. By gathering feedback through this structured lens, we aimed to carve out a roadmap for the subsequent development and refinement of x-[plAIn] that we plan to offer via the \href{https://openai.com/blog/introducing-the-gpt-store}{GPT Store}.

\begin{figure*}[ht!]
\centering
\begin{subfigure}[b]{0.4\linewidth}
  \centering
  \includegraphics[width=\linewidth]{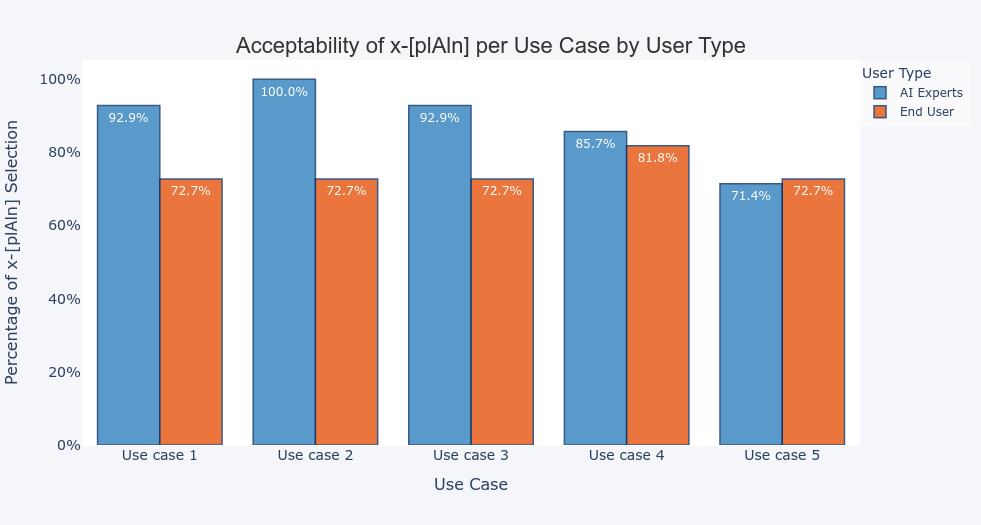}
  \caption{Acceptability of x-[plAIn] concerning role of the users.}
  \label{fig:xplain_role}
\end{subfigure}
\hspace{0.1\linewidth} 
\begin{subfigure}[b]{0.4\linewidth}
  \centering
  \includegraphics[width=\linewidth]{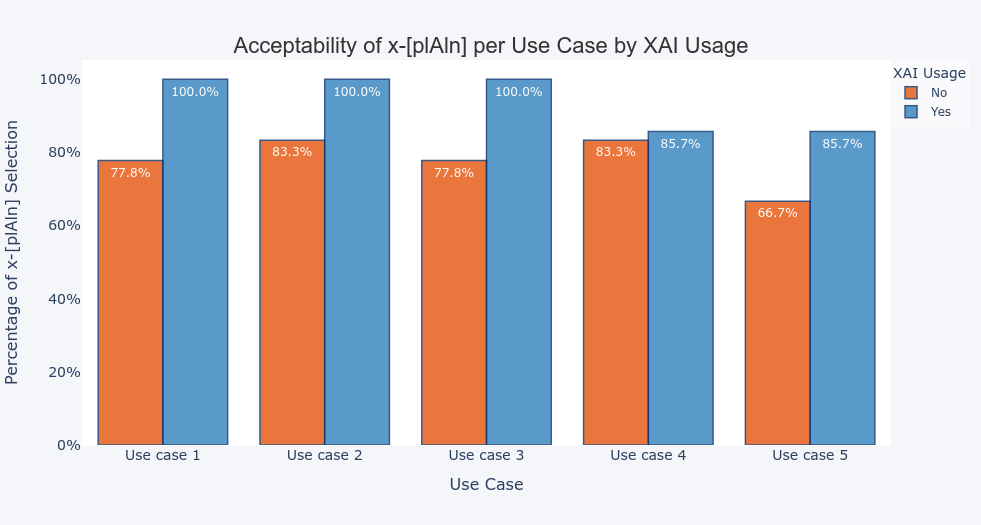}
  \caption{Acceptability of x-[plAIn] concerning the usage of XAI of the users.}
  \label{fig:xplain_xai}
\end{subfigure}
\caption{Comparison of Acceptability}
\label{fig:acceptability_comparison}
\end{figure*}

\begin{figure*}[ht!]
\centering
\includegraphics[width=0.5\linewidth]{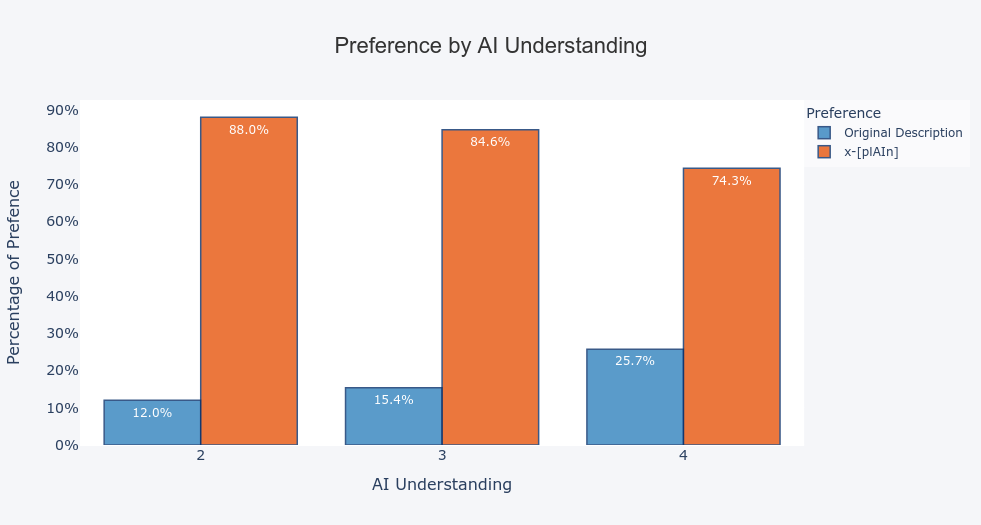}
\caption{Description preference based on perceived AI understanding level.}
\label{fig:xplain_aiunderstanding}
\end{figure*}
Through the conducted questionnaire, it was revealed that a significant majority of participants, exceeding 70\%, expressed a satisfaction level below 60\% concerning their comprehension of AI-based decision models. This is further underscored by the fact that a mere 30\% of participants are actively employing XAI methodologies. This finding raises critical questions about the prevalence and perceived efficacy of XAI techniques within the industry. When it comes to scenario-based preferences, over 80\% of participants favored x-[plAIn] descriptions in comparison to the conventional descriptions derived from original papers associated with XAI methods, particularly in decision-making contexts and image comprehension.

The feedback on enhancing x-[plAIn] predominantly revolved around the brevity of responses, given the tendency of GPT models towards verbosity, and the need for tailoring explanations to suit the specific background or domain of the end-user. While the model inherently possesses the capability to customize responses based on domain-specific information, this feature was not fully showcased to the participants due to the absence of domain information in the prompts, which was intentionally omitted to ensure a level playing field in the assessment. This suggests a potential area for refinement in future iterations, where the model's adaptive response generation can be demonstrated more effectively to respondents.

The comparative analysis of the two bar plots reveals insightful trends about the acceptability of "x-[plAIn]" across different user groups and their engagement with XAI methods.

In Figure \ref{fig:xplain_role}, which distinguishes between end users and AI experts, a distinct pattern emerges. End users, presumably less versed in the technical aspects of AI, show varying levels of preference for "x-[plAIn]" across different use cases. This variability could indicate a nuanced approach to AI explanations, where the complexity or context of each use case significantly influences their preference. On the other hand, AI experts, with their deeper technical understanding, exhibit a more consistent response pattern across the use cases. Their preferences might reflect a critical evaluation of "x-[plAIn]" against their advanced understanding of AI processes.

Figure \ref{fig:xplain_xai}, focuses on the usage of XAI methods and also provides compelling insights. Respondents who actively use XAI methods demonstrate a certain level of preference for "x-[plAIn]", which might suggest that their familiarity with XAI influences their expectations and acceptance of explanatory tools. Conversely, non-XAI users, who might not have a benchmark for comparing such tools, show differing degrees of acceptance for "x-[plAIn]", potentially guided more by the tool's clarity and usability than by its technical robustness.

Figure \ref{fig:xplain_aiunderstanding} demonstrates a notable trend concerning users' preferences in relation to their self-reported comprehension of AI model outputs. This graphical representation indicates a discernible correlation between the level of claimed understanding and the preferences exhibited by respondents. It appears that individuals professing a more profound grasp of AI model outputs tend to require less information, potentially influencing a shift in their preferences. Despite this observed trend, it is noteworthy that x-[plAIn] retains a significant degree of favorability, being the preferred choice in 75\% of instances among the cohort exhibiting the highest level of understanding.

\section{Conclusion}

For future enhancements, it will be crucial to implement features that allow end-users to specify (more strictly) their preference for the level of detail they require. A verbose setting could cater to those looking for in-depth understanding, while a more streamlined option would benefit users seeking brief clarifications. Such a choice empowers users, providing a user-centric approach that accommodates a wide range of use cases from novice inquiries to expert validations.

Additionally, the feedback highlights the importance of considering user experience, particularly for those unfamiliar with the subject matter. Breaking down complex topics into smaller, individually explained segments can significantly enhance comprehension. Conversely, for experienced users, lengthy and information-dense responses may prove unnecessary and time-consuming. To this end, introducing an option to toggle between longer and shorter answer formats while shifting focus from understanding the XAI methods to the extraction of insights can be beneficial.

In future work, we aim to investigate a specific characteristic of this tool, identified during the development of x-[plAIn]. This tool holds potential for experienced AI engineers, offering a resource to pinpoint and mitigate potential biases that may emerge at different phases of the model creation pipeline. These phases include data collection, preprocessing, model training, and validation processes. By utilizing this tool, AI professionals can significantly contribute to the cultivation of AI systems that excel not only in technical prowess but also in ethical integrity and social responsibility.

\section*{Acknowledgements}
The research leading to the results presented in this paper has received funding from the Europeans Union’s funded Project HumAIne under grant agreement no 101120218.




\bibliographystyle{elsarticle-harv} 
\bibliography{example}

\begin{thebibliography}{45}
\expandafter\ifx\csname natexlab\endcsname\relax\def\natexlab#1{#1}\fi
\providecommand{\url}[1]{\texttt{#1}}
\providecommand{\href}[2]{#2}
\providecommand{\path}[1]{#1}
\providecommand{\DOIprefix}{doi:}
\providecommand{\ArXivprefix}{arXiv:}
\providecommand{\URLprefix}{URL: }
\providecommand{\Pubmedprefix}{pmid:}
\providecommand{\doi}[1]{\href{http://dx.doi.org/#1}{\path{#1}}}
\providecommand{\Pubmed}[1]{\href{pmid:#1}{\path{#1}}}
\providecommand{\bibinfo}[2]{#2}
\ifx\xfnm\relax \def\xfnm[#1]{\unskip,\space#1}\fi
\bibitem[{Ali and Kostakos(2023)}]{ali2023huntgpt}
\bibinfo{author}{Ali, T.}, \bibinfo{author}{Kostakos, P.}, \bibinfo{year}{2023}.
\newblock \bibinfo{title}{Huntgpt: Integrating machine learning-based anomaly detection and explainable ai with large language models (llms)}.
\newblock \bibinfo{journal}{arXiv preprint arXiv:2309.16021} .
\bibitem[{Arrieta et~al.(2020)Arrieta, D{\'\i}az-Rodr{\'\i}guez, Del~Ser, Bennetot, Tabik, Barbado, Garc{\'\i}a, Gil-L{\'o}pez, Molina, Benjamins et~al.}]{arrieta2020explainable}
\bibinfo{author}{Arrieta, A.B.}, \bibinfo{author}{D{\'\i}az-Rodr{\'\i}guez, N.}, \bibinfo{author}{Del~Ser, J.}, \bibinfo{author}{Bennetot, A.}, \bibinfo{author}{Tabik, S.}, \bibinfo{author}{Barbado, A.}, \bibinfo{author}{Garc{\'\i}a, S.}, \bibinfo{author}{Gil-L{\'o}pez, S.}, \bibinfo{author}{Molina, D.}, \bibinfo{author}{Benjamins, R.}, et~al., \bibinfo{year}{2020}.
\newblock \bibinfo{title}{Explainable artificial intelligence (xai): Concepts, taxonomies, opportunities and challenges toward responsible ai}.
\newblock \bibinfo{journal}{Information fusion} \bibinfo{volume}{58}, \bibinfo{pages}{82--115}.
\bibitem[{Baxter(2018)}]{baxter2018meet}
\bibinfo{author}{Baxter, K.}, \bibinfo{year}{2018}.
\newblock \bibinfo{title}{How to meet user expectations for artifcial intelligence}.
\newblock \bibinfo{journal}{Medium. Retrieved September} .
\bibitem[{Brown et~al.(2020)Brown, Mann, Ryder, Subbiah, Kaplan, Dhariwal, Neelakantan, Shyam, Sastry, Askell et~al.}]{brown2020language}
\bibinfo{author}{Brown, T.}, \bibinfo{author}{Mann, B.}, \bibinfo{author}{Ryder, N.}, \bibinfo{author}{Subbiah, M.}, \bibinfo{author}{Kaplan, J.D.}, \bibinfo{author}{Dhariwal, P.}, \bibinfo{author}{Neelakantan, A.}, \bibinfo{author}{Shyam, P.}, \bibinfo{author}{Sastry, G.}, \bibinfo{author}{Askell, A.}, et~al., \bibinfo{year}{2020}.
\newblock \bibinfo{title}{Language models are few-shot learners}.
\newblock \bibinfo{journal}{Advances in neural information processing systems} \bibinfo{volume}{33}, \bibinfo{pages}{1877--1901}.
\bibitem[{Chakraborty et~al.(2017)Chakraborty, Tomsett, Raghavendra, Harborne, Alzantot, Cerutti, Srivastava, Preece, Julier, Rao et~al.}]{chakraborty2017interpretability}
\bibinfo{author}{Chakraborty, S.}, \bibinfo{author}{Tomsett, R.}, \bibinfo{author}{Raghavendra, R.}, \bibinfo{author}{Harborne, D.}, \bibinfo{author}{Alzantot, M.}, \bibinfo{author}{Cerutti, F.}, \bibinfo{author}{Srivastava, M.}, \bibinfo{author}{Preece, A.}, \bibinfo{author}{Julier, S.}, \bibinfo{author}{Rao, R.M.}, et~al., \bibinfo{year}{2017}.
\newblock \bibinfo{title}{Interpretability of deep learning models: A survey of results}, in: \bibinfo{booktitle}{2017 IEEE smartworld, ubiquitous intelligence \& computing, advanced \& trusted computed, scalable computing \& communications, cloud \& big data computing, Internet of people and smart city innovation (smartworld/SCALCOM/UIC/ATC/CBDcom/IOP/SCI)}, \bibinfo{organization}{IEEE}. pp. \bibinfo{pages}{1--6}.
\bibitem[{Choo and Liu(2018)}]{choo2018visual}
\bibinfo{author}{Choo, J.}, \bibinfo{author}{Liu, S.}, \bibinfo{year}{2018}.
\newblock \bibinfo{title}{Visual analytics for explainable deep learning}.
\newblock \bibinfo{journal}{IEEE computer graphics and applications} \bibinfo{volume}{38}, \bibinfo{pages}{84--92}.
\bibitem[{Chun and Elkins(2023)}]{chun2023explainable}
\bibinfo{author}{Chun, J.}, \bibinfo{author}{Elkins, K.}, \bibinfo{year}{2023}.
\newblock \bibinfo{title}{explainable ai with gpt4 for story analysis and generation: A novel framework for diachronic sentiment analysis}.
\newblock \bibinfo{journal}{International Journal of Digital Humanities} , \bibinfo{pages}{1--26}.
\bibitem[{Dakhel et~al.(2023)Dakhel, Majdinasab, Nikanjam, Khomh, Desmarais and Jiang}]{dakhel2023github}
\bibinfo{author}{Dakhel, A.M.}, \bibinfo{author}{Majdinasab, V.}, \bibinfo{author}{Nikanjam, A.}, \bibinfo{author}{Khomh, F.}, \bibinfo{author}{Desmarais, M.C.}, \bibinfo{author}{Jiang, Z.M.}, \bibinfo{year}{2023}.
\newblock \bibinfo{title}{Github copilot ai pair programmer: Asset or liability?}
\newblock \bibinfo{journal}{Journal of Systems and Software} , \bibinfo{pages}{111734}.
\bibitem[{Devlin et~al.(2018)Devlin, Chang, Lee and Toutanova}]{devlin2018bert}
\bibinfo{author}{Devlin, J.}, \bibinfo{author}{Chang, M.W.}, \bibinfo{author}{Lee, K.}, \bibinfo{author}{Toutanova, K.}, \bibinfo{year}{2018}.
\newblock \bibinfo{title}{Bert: Pre-training of deep bidirectional transformers for language understanding}.
\newblock \bibinfo{journal}{arXiv preprint arXiv:1810.04805} .
\bibitem[{Fatouros et~al.(2024)Fatouros, Metaxas, Soldatos and Kyriazis}]{fatouros2024can}
\bibinfo{author}{Fatouros, G.}, \bibinfo{author}{Metaxas, K.}, \bibinfo{author}{Soldatos, J.}, \bibinfo{author}{Kyriazis, D.}, \bibinfo{year}{2024}.
\newblock \bibinfo{title}{Can large language models beat wall street? unveiling the potential of ai in stock selection}.
\newblock \bibinfo{journal}{arXiv preprint arXiv:2401.03737} .
\bibitem[{Fatouros et~al.(2023)Fatouros, Soldatos, Kouroumali, Makridis and Kyriazis}]{fatouros2023transforming}
\bibinfo{author}{Fatouros, G.}, \bibinfo{author}{Soldatos, J.}, \bibinfo{author}{Kouroumali, K.}, \bibinfo{author}{Makridis, G.}, \bibinfo{author}{Kyriazis, D.}, \bibinfo{year}{2023}.
\newblock \bibinfo{title}{Transforming sentiment analysis in the financial domain with chatgpt}.
\newblock \bibinfo{journal}{Machine Learning with Applications} \bibinfo{volume}{14}, \bibinfo{pages}{100508}.
\bibitem[{Feldhus et~al.(2023)Feldhus, Hennig, Nasert, Ebert, Schwarzenberg and Mller}]{feldhus2023saliency}
\bibinfo{author}{Feldhus, N.}, \bibinfo{author}{Hennig, L.}, \bibinfo{author}{Nasert, M.}, \bibinfo{author}{Ebert, C.}, \bibinfo{author}{Schwarzenberg, R.}, \bibinfo{author}{Mller, S.}, \bibinfo{year}{2023}.
\newblock \bibinfo{title}{Saliency map verbalization: Comparing feature importance representations from model-free and instruction-based methods}, in: \bibinfo{booktitle}{Proceedings of the 1st Workshop on Natural Language Reasoning and Structured Explanations (NLRSE)}, pp. \bibinfo{pages}{30--46}.
\bibitem[{Gramegna and Giudici(2021)}]{gramegna2021shap}
\bibinfo{author}{Gramegna, A.}, \bibinfo{author}{Giudici, P.}, \bibinfo{year}{2021}.
\newblock \bibinfo{title}{Shap and lime: an evaluation of discriminative power in credit risk}.
\newblock \bibinfo{journal}{Frontiers in Artificial Intelligence} \bibinfo{volume}{4}, \bibinfo{pages}{752558}.
\bibitem[{Gunning(2016)}]{gunning2016explainable}
\bibinfo{author}{Gunning, D.}, \bibinfo{year}{2016}.
\newblock \bibinfo{title}{Explainable artificial intelligence (xai) darpa-baa-16-53}.
\newblock \bibinfo{journal}{Defense Advanced Research Projects Agency} .
\bibitem[{Jan et~al.(2020)Jan, Ishakian and Muthusamy}]{jan2020ai}
\bibinfo{author}{Jan, S.T.}, \bibinfo{author}{Ishakian, V.}, \bibinfo{author}{Muthusamy, V.}, \bibinfo{year}{2020}.
\newblock \bibinfo{title}{Ai trust in business processes: the need for process-aware explanations}, in: \bibinfo{booktitle}{Proceedings of the AAAI Conference on Artificial Intelligence}, pp. \bibinfo{pages}{13403--13404}.
\bibitem[{JasperAI(2023)}]{jasper23report}
\bibinfo{author}{JasperAI}, \bibinfo{year}{2023}.
\newblock \bibinfo{title}{The ai in business trend report}.
\newblock \URLprefix \url{https://www.jasper.ai/blog/ai-business-trend-report}. \bibinfo{note}{accessed:May 26, 2023}.
\bibitem[{Klein(2020)}]{klein2020reducing}
\bibinfo{author}{Klein, A.}, \bibinfo{year}{2020}.
\newblock \bibinfo{title}{Reducing bias in ai-based financial services} .
\bibitem[{Lepenioti et~al.(2020)Lepenioti, Bousdekis, Apostolou and Mentzas}]{lepenioti2020prescriptive}
\bibinfo{author}{Lepenioti, K.}, \bibinfo{author}{Bousdekis, A.}, \bibinfo{author}{Apostolou, D.}, \bibinfo{author}{Mentzas, G.}, \bibinfo{year}{2020}.
\newblock \bibinfo{title}{Prescriptive analytics: Literature review and research challenges}.
\newblock \bibinfo{journal}{International Journal of Information Management} \bibinfo{volume}{50}, \bibinfo{pages}{57--70}.
\bibitem[{Lim and Dey(2009)}]{lim2009assessing}
\bibinfo{author}{Lim, B.Y.}, \bibinfo{author}{Dey, A.K.}, \bibinfo{year}{2009}.
\newblock \bibinfo{title}{Assessing demand for intelligibility in context-aware applications}, in: \bibinfo{booktitle}{Proceedings of the 11th international conference on Ubiquitous computing}, pp. \bibinfo{pages}{195--204}.
\bibitem[{Lundberg and Lee(2017)}]{lundberg2017unified}
\bibinfo{author}{Lundberg, S.M.}, \bibinfo{author}{Lee, S.I.}, \bibinfo{year}{2017}.
\newblock \bibinfo{title}{A unified approach to interpreting model predictions}.
\newblock \bibinfo{journal}{Advances in neural information processing systems} \bibinfo{volume}{30}.
\bibitem[{Makridis et~al.(2023a)Makridis, Fatouros, Kiourtis, Kotios, Koukos, Kyriazis and Soldatos}]{makridis2023towards}
\bibinfo{author}{Makridis, G.}, \bibinfo{author}{Fatouros, G.}, \bibinfo{author}{Kiourtis, A.}, \bibinfo{author}{Kotios, D.}, \bibinfo{author}{Koukos, V.}, \bibinfo{author}{Kyriazis, D.}, \bibinfo{author}{Soldatos, J.}, \bibinfo{year}{2023}a.
\newblock \bibinfo{title}{Towards a unified multidimensional explainability metric: Evaluating trustworthiness in ai models}, in: \bibinfo{booktitle}{2023 19th International Conference on Distributed Computing in Smart Systems and the Internet of Things (DCOSS-IoT)}, \bibinfo{organization}{IEEE}. pp. \bibinfo{pages}{504--511}.
\bibitem[{Makridis et~al.(2023b)Makridis, Fatouros, Koukos, Kotios, Kyriazis and Soldatos}]{makridis2023xai}
\bibinfo{author}{Makridis, G.}, \bibinfo{author}{Fatouros, G.}, \bibinfo{author}{Koukos, V.}, \bibinfo{author}{Kotios, D.}, \bibinfo{author}{Kyriazis, D.}, \bibinfo{author}{Soldatos, I.}, \bibinfo{year}{2023}b.
\newblock \bibinfo{title}{Xai for time-series classification leveraging image highlight methods}.
\newblock \bibinfo{journal}{arXiv preprint arXiv:2311.17110} .
\bibitem[{Makridis et~al.(2022)Makridis, Heyrman, Kotios, Mavrepis, Callens, Van De~Vijver, Maselyne, Aluw{\'e} and Kyriazis}]{makridis2022evaluating}
\bibinfo{author}{Makridis, G.}, \bibinfo{author}{Heyrman, E.}, \bibinfo{author}{Kotios, D.}, \bibinfo{author}{Mavrepis, P.}, \bibinfo{author}{Callens, B.}, \bibinfo{author}{Van De~Vijver, R.}, \bibinfo{author}{Maselyne, J.}, \bibinfo{author}{Aluw{\'e}, M.}, \bibinfo{author}{Kyriazis, D.}, \bibinfo{year}{2022}.
\newblock \bibinfo{title}{Evaluating machine learning techniques to define the factors related to boar taint}.
\newblock \bibinfo{journal}{Livestock Science} \bibinfo{volume}{264}, \bibinfo{pages}{105045}.
\bibitem[{Makridis et~al.(2020)Makridis, Kyriazis and Plitsos}]{makridis2020predictive}
\bibinfo{author}{Makridis, G.}, \bibinfo{author}{Kyriazis, D.}, \bibinfo{author}{Plitsos, S.}, \bibinfo{year}{2020}.
\newblock \bibinfo{title}{Predictive maintenance leveraging machine learning for time-series forecasting in the maritime industry}, in: \bibinfo{booktitle}{2020 IEEE 23rd International Conference on Intelligent Transportation Systems (ITSC)}, \bibinfo{organization}{IEEE}. pp. \bibinfo{pages}{1--8}.
\bibitem[{Molnar et~al.(2020)Molnar, Casalicchio and Bischl}]{molnar2020interpretable}
\bibinfo{author}{Molnar, C.}, \bibinfo{author}{Casalicchio, G.}, \bibinfo{author}{Bischl, B.}, \bibinfo{year}{2020}.
\newblock \bibinfo{title}{Interpretable machine learning--a brief history, state-of-the-art and challenges}, in: \bibinfo{booktitle}{Joint European conference on machine learning and knowledge discovery in databases}, \bibinfo{organization}{Springer}. pp. \bibinfo{pages}{417--431}.
\bibitem[{Moosbauer et~al.(2021)Moosbauer, Herbinger, Casalicchio, Lindauer and Bischl}]{moosbauer2021explaining}
\bibinfo{author}{Moosbauer, J.}, \bibinfo{author}{Herbinger, J.}, \bibinfo{author}{Casalicchio, G.}, \bibinfo{author}{Lindauer, M.}, \bibinfo{author}{Bischl, B.}, \bibinfo{year}{2021}.
\newblock \bibinfo{title}{Explaining hyperparameter optimization via partial dependence plots}.
\newblock \bibinfo{journal}{Advances in Neural Information Processing Systems} \bibinfo{volume}{34}, \bibinfo{pages}{2280--2291}.
\bibitem[{Moujahid et~al.(2022)Moujahid, Cherradi, Al-Sarem, Bahatti, Eljialy, Alsaeedi and Saeed}]{moujahid2022combining}
\bibinfo{author}{Moujahid, H.}, \bibinfo{author}{Cherradi, B.}, \bibinfo{author}{Al-Sarem, M.}, \bibinfo{author}{Bahatti, L.}, \bibinfo{author}{Eljialy, A.B.A.M.Y.}, \bibinfo{author}{Alsaeedi, A.}, \bibinfo{author}{Saeed, F.}, \bibinfo{year}{2022}.
\newblock \bibinfo{title}{Combining cnn and grad-cam for covid-19 disease prediction and visual explanation.}
\newblock \bibinfo{journal}{Intelligent Automation \& Soft Computing} \bibinfo{volume}{32}.
\bibitem[{Ribeiro et~al.(2016)Ribeiro, Singh and Guestrin}]{ribeiro2016should}
\bibinfo{author}{Ribeiro, M.T.}, \bibinfo{author}{Singh, S.}, \bibinfo{author}{Guestrin, C.}, \bibinfo{year}{2016}.
\newblock \bibinfo{title}{" why should i trust you?" explaining the predictions of any classifier}, in: \bibinfo{booktitle}{Proceedings of the 22nd ACM SIGKDD international conference on knowledge discovery and data mining}, pp. \bibinfo{pages}{1135--1144}.
\bibitem[{Ribeiro et~al.(2018)Ribeiro, Singh and Guestrin}]{ribeiro2018anchors}
\bibinfo{author}{Ribeiro, M.T.}, \bibinfo{author}{Singh, S.}, \bibinfo{author}{Guestrin, C.}, \bibinfo{year}{2018}.
\newblock \bibinfo{title}{Anchors: High-precision model-agnostic explanations}, in: \bibinfo{booktitle}{Proceedings of the AAAI conference on artificial intelligence}.
\bibitem[{Saeed and Omlin(2023)}]{saeed2023explainable}
\bibinfo{author}{Saeed, W.}, \bibinfo{author}{Omlin, C.}, \bibinfo{year}{2023}.
\newblock \bibinfo{title}{Explainable ai (xai): A systematic meta-survey of current challenges and future opportunities}.
\newblock \bibinfo{journal}{Knowledge-Based Systems} \bibinfo{volume}{263}, \bibinfo{pages}{110273}.
\bibitem[{Sallam(2023)}]{sallam2023chatgpt}
\bibinfo{author}{Sallam, M.}, \bibinfo{year}{2023}.
\newblock \bibinfo{title}{Chatgpt utility in healthcare education, research, and practice: systematic review on the promising perspectives and valid concerns}, in: \bibinfo{booktitle}{Healthcare}, \bibinfo{organization}{MDPI}. p. \bibinfo{pages}{887}.
\bibitem[{Scao et~al.(2022)Scao, Fan, Akiki, Pavlick, Ili{\'c}, Hesslow, Castagn{\'e}, Luccioni, Yvon, Gall{\'e} et~al.}]{scao2022bloom}
\bibinfo{author}{Scao, T.L.}, \bibinfo{author}{Fan, A.}, \bibinfo{author}{Akiki, C.}, \bibinfo{author}{Pavlick, E.}, \bibinfo{author}{Ili{\'c}, S.}, \bibinfo{author}{Hesslow, D.}, \bibinfo{author}{Castagn{\'e}, R.}, \bibinfo{author}{Luccioni, A.S.}, \bibinfo{author}{Yvon, F.}, \bibinfo{author}{Gall{\'e}, M.}, et~al., \bibinfo{year}{2022}.
\newblock \bibinfo{title}{Bloom: A 176b-parameter open-access multilingual language model}.
\newblock \bibinfo{journal}{arXiv preprint arXiv:2211.05100} .
\bibitem[{Selvaraju et~al.(2017)Selvaraju, Cogswell, Das, Vedantam, Parikh and Batra}]{selvaraju2017grad}
\bibinfo{author}{Selvaraju, R.R.}, \bibinfo{author}{Cogswell, M.}, \bibinfo{author}{Das, A.}, \bibinfo{author}{Vedantam, R.}, \bibinfo{author}{Parikh, D.}, \bibinfo{author}{Batra, D.}, \bibinfo{year}{2017}.
\newblock \bibinfo{title}{Grad-cam: Visual explanations from deep networks via gradient-based localization}, in: \bibinfo{booktitle}{Proceedings of the IEEE international conference on computer vision}, pp. \bibinfo{pages}{618--626}.
\bibitem[{Soldatos and Kyriazis(2021)}]{soldatos2021trusted}
\bibinfo{author}{Soldatos, J.}, \bibinfo{author}{Kyriazis, D.}, \bibinfo{year}{2021}.
\newblock \bibinfo{title}{Trusted artificial intelligence in manufacturing; trusted artificial intelligence in manufacturing: A review of the emerging wave of ethical and human centric ai technologies for smart production; a review of the emerging wave of ethical and human centric ai technologies for smart production} .
\bibitem[{Stolfo et~al.(1999)Stolfo, Fan, Lee, Prodromidis and Chan}]{misckddcup1999data130}
\bibinfo{author}{Stolfo, S.}, \bibinfo{author}{Fan, W.}, \bibinfo{author}{Lee, W.}, \bibinfo{author}{Prodromidis, A.}, \bibinfo{author}{Chan, P.}, \bibinfo{year}{1999}.
\newblock \bibinfo{title}{Kdd cup 1999 data}.
\newblock \bibinfo{howpublished}{UCI Machine Learning Repository}.
\newblock \bibinfo{note}{DOI: https://doi.org/10.24432/C51C7N}.
\bibitem[{Susnjak(2023)}]{susnjak2023beyond}
\bibinfo{author}{Susnjak, T.}, \bibinfo{year}{2023}.
\newblock \bibinfo{title}{Beyond predictive learning analytics modelling and onto explainable artificial intelligence with prescriptive analytics and chatgpt}.
\newblock \bibinfo{journal}{International Journal of Artificial Intelligence in Education} , \bibinfo{pages}{1--31}.
\bibitem[{Szczepa{\'n}ski et~al.(2021)Szczepa{\'n}ski, Pawlicki, Kozik and Chora{\'s}}]{szczepanski2021new}
\bibinfo{author}{Szczepa{\'n}ski, M.}, \bibinfo{author}{Pawlicki, M.}, \bibinfo{author}{Kozik, R.}, \bibinfo{author}{Chora{\'s}, M.}, \bibinfo{year}{2021}.
\newblock \bibinfo{title}{New explainability method for bert-based model in fake news detection}.
\newblock \bibinfo{journal}{Scientific reports} \bibinfo{volume}{11}, \bibinfo{pages}{23705}.
\bibitem[{Tan(2023)}]{tan2023causal}
\bibinfo{author}{Tan, J.T.}, \bibinfo{year}{2023}.
\newblock \bibinfo{title}{Causal abstraction for chain-of-thought reasoning in arithmetic word problems}, in: \bibinfo{booktitle}{Proceedings of the 6th BlackboxNLP Workshop: Analyzing and Interpreting Neural Networks for NLP}, pp. \bibinfo{pages}{155--168}.
\bibitem[{Thoppilan et~al.(2022)Thoppilan, De~Freitas, Hall, Shazeer, Kulshreshtha, Cheng, Jin, Bos, Baker, Du et~al.}]{thoppilan2022lamda}
\bibinfo{author}{Thoppilan, R.}, \bibinfo{author}{De~Freitas, D.}, \bibinfo{author}{Hall, J.}, \bibinfo{author}{Shazeer, N.}, \bibinfo{author}{Kulshreshtha, A.}, \bibinfo{author}{Cheng, H.T.}, \bibinfo{author}{Jin, A.}, \bibinfo{author}{Bos, T.}, \bibinfo{author}{Baker, L.}, \bibinfo{author}{Du, Y.}, et~al., \bibinfo{year}{2022}.
\newblock \bibinfo{title}{Lamda: Language models for dialog applications}.
\newblock \bibinfo{journal}{arXiv preprint arXiv:2201.08239} .
\bibitem[{Torralba and Efros(2011)}]{torralba2011unbiased}
\bibinfo{author}{Torralba, A.}, \bibinfo{author}{Efros, A.A.}, \bibinfo{year}{2011}.
\newblock \bibinfo{title}{Unbiased look at dataset bias}, in: \bibinfo{booktitle}{CVPR 2011}, \bibinfo{organization}{IEEE}. pp. \bibinfo{pages}{1521--1528}.
\bibitem[{Virtue(2017)}]{virtue2017designing}
\bibinfo{author}{Virtue, E.}, \bibinfo{year}{2017}.
\newblock \bibinfo{title}{Designing with ai}.
\newblock \bibinfo{journal}{Retrieved July} \bibinfo{volume}{29}, \bibinfo{pages}{2022}.
\bibitem[{Wachter et~al.(2017)Wachter, Mittelstadt and Russell}]{wachter2017counterfactual}
\bibinfo{author}{Wachter, S.}, \bibinfo{author}{Mittelstadt, B.}, \bibinfo{author}{Russell, C.}, \bibinfo{year}{2017}.
\newblock \bibinfo{title}{Counterfactual explanations without opening the black box: Automated decisions and the gdpr}.
\newblock \bibinfo{journal}{Harv. JL \& Tech.} \bibinfo{volume}{31}, \bibinfo{pages}{841}.
\bibitem[{Wei et~al.(2022)Wei, Wang, Schuurmans, Bosma, Xia, Chi, Le, Zhou et~al.}]{wei2022chain}
\bibinfo{author}{Wei, J.}, \bibinfo{author}{Wang, X.}, \bibinfo{author}{Schuurmans, D.}, \bibinfo{author}{Bosma, M.}, \bibinfo{author}{Xia, F.}, \bibinfo{author}{Chi, E.}, \bibinfo{author}{Le, Q.V.}, \bibinfo{author}{Zhou, D.}, et~al., \bibinfo{year}{2022}.
\newblock \bibinfo{title}{Chain-of-thought prompting elicits reasoning in large language models}.
\newblock \bibinfo{journal}{Advances in Neural Information Processing Systems} \bibinfo{volume}{35}, \bibinfo{pages}{24824--24837}.
\bibitem[{Wu et~al.(2023)Wu, Irsoy, Lu, Dabravolski, Dredze, Gehrmann, Kambadur, Rosenberg and Mann}]{wu2023bloomberggpt}
\bibinfo{author}{Wu, S.}, \bibinfo{author}{Irsoy, O.}, \bibinfo{author}{Lu, S.}, \bibinfo{author}{Dabravolski, V.}, \bibinfo{author}{Dredze, M.}, \bibinfo{author}{Gehrmann, S.}, \bibinfo{author}{Kambadur, P.}, \bibinfo{author}{Rosenberg, D.}, \bibinfo{author}{Mann, G.}, \bibinfo{year}{2023}.
\newblock \bibinfo{title}{Bloomberggpt: A large language model for finance}.
\newblock \bibinfo{journal}{arXiv preprint arXiv:2303.17564} .
\bibitem[{Zahavy et~al.(2016)Zahavy, Ben-Zrihem and Mannor}]{zahavy2016graying}
\bibinfo{author}{Zahavy, T.}, \bibinfo{author}{Ben-Zrihem, N.}, \bibinfo{author}{Mannor, S.}, \bibinfo{year}{2016}.
\newblock \bibinfo{title}{Graying the black box: Understanding dqns}, in: \bibinfo{booktitle}{International conference on machine learning}, \bibinfo{organization}{PMLR}. pp. \bibinfo{pages}{1899--1908}.

\end{thebibliography}






\end{document}